\documentclass[conference]{IEEEtran}
\IEEEoverridecommandlockouts
\usepackage{cite}
\usepackage{amsmath,amssymb,amsfonts}
\usepackage[noend]{algorithmic}
\usepackage{graphicx}
\usepackage{textcomp}
\usepackage{xcolor}
\usepackage{booktabs}
\usepackage{multirow}
\usepackage{microtype}
\usepackage{hyperref}
\usepackage{cleveref}
\usepackage{todonotes}
\usepackage{algorithm}
\usepackage{subcaption}
\usepackage{pgfplots}
\pgfplotsset{compat=1.17}

\def\BibTeX{{\rm B\kern-.05em{\sc i\kern-.025em b}\kern-.08em
    T\kern-.1667em\lower.7ex\hbox{E}\kern-.125emX}}
\begin{document}

\IEEEoverridecommandlockouts
\IEEEpubid{\makebox[\columnwidth]{ 979-8-3503-5067-8/24/\$31.00~\copyright2024 IEEE \hfill} 
\hspace{\columnsep}\makebox[\columnwidth]{ }}

\title{Match Point AI: A Novel AI Framework for Evaluating
Data-Driven Tennis Strategies}


\author{
    \IEEEauthorblockN{Carlo Nübel}
    \IEEEauthorblockA{\textit{Faculty of Computer Science} \\
        \textit{Otto-von-Guericke-University Magdeburg}\\
        Magdeburg, Germany \\
        carlo.nuebel@ovgu.de} 
    \and
    \IEEEauthorblockN{Alexander Dockhorn}
    \IEEEauthorblockA{\textit{Faculty of EECS} \\
    \textit{Leibniz University Hannover}\\
    Hannover, Germany \\
    dockhorn@tnt.uni-hannover.de} 
    \and
    \IEEEauthorblockN{Sanaz Mostaghim}
    \IEEEauthorblockA{\textit{Faculty of Computer Science} \\
        \textit{Otto-von-Guericke-University Magdeburg}\\
        Magdeburg, Germany \\
        \textit{Fraunhofer Institute for Transportation} \\ 
        \textit{and Infrastructure Systems IVI}\\
        Dresden, Germany \\
        sanaz.mostaghim@ovgu.de}
}

\maketitle

\begin{abstract}
Many works in the domain of artificial intelligence in games focus on board or video games due to the ease of reimplementing their mechanics~\cite{DocGru2020,GaiBal2020}.
Decision-making problems in real-world sports share many similarities to such domains.
Nevertheless, not many frameworks on sports games exist. 
In this paper, we present the tennis match simulation environment \textit{Match Point AI}, in which different agents can compete against real-world data-driven bot strategies.
Next to presenting the framework, we highlight its capabilities by illustrating, how MCTS can be used in Match Point AI to optimize the shot direction selection problem in tennis.
While the framework will be extended in the future, first experiments already reveal that generated shot-by-shot data of simulated tennis matches show realistic characteristics when compared to real-world data. 
At the same time, reasonable shot placement strategies emerge, which share similarities to the ones found in real-world tennis matches. 
\end{abstract}



\section{Introduction}

For two decades, the men's tennis scene has revolved around the dominant trio dubbed "the Big Three". From 2003 to 2023, they claimed 66 out of 80 Grand Slam titles. While Federer retired and Nadal battled injuries, Djokovic, the last of the triumvirate remains unmatched. This paper explores AI's potential in tennis decision-making, aiming to dissect strategies, enhance shot placements, and potentially aid emerging players in challenging Djokovic's reign.

Sports data analytics has been shown to provide deep insights into athletic performance and tactical patterns. While analyzing real-world data can already help us to revise and improve our knowledge of a sport in general or an upcoming opponent, it does not allow us to test new strategies and how they would perform against others. Artificial intelligence can bridge this gap, by running simulations in which the behavior of our opponent is imitated, allowing us to specifically train against it. Hence, we developed the Match Point AI framework, which models Tennis matches and players based on historical data and aims to empower AI agents in optimizing tennis strategies.

In sports, particularly tennis, Monte Carlo Tree Search (MCTS) has seen limited application so far. For instance, \cite{mcts_in_tennis} implemented MCTS in a 3D tennis video game to develop adjustable, believable AI behavior. Additionally, an observation-based AI method was employed in a mobile tennis game to simulate human-like behavior \cite{tennis_in_CIG}. In contrast to these studies, our research focuses on how our framework can be used to identify the most effective shot direction selection strategies in a tennis rally, based on real-world data, when competing against actual tennis strategies.

In this paper, we first give a brief overview of Tennis (\cref{sec:tennis}). Further, we are presenting the Match Point AI framework and the concepts implemented to yield an accurate representation of play-styles from professional Tennis players \cref{sec:framework}. \cref{sec:experiments} highlights the insights gained from running simulations with the proposed framework and different variations of MCTS. The resulting data of those experiments is then analyzed to see which strategies and shot patterns are identified by the MCTS algorithm and to gain insight into which adaptation is best suited to solve the decision-making problems in Tennis. We conclude the paper by providing an overview of future extensions of the framework and their potential benefits.


\section{Tennis - Rules and Terminology}
\label{sec:tennis}

For a better understanding of the presented framework, we provide a short introduction to tennis and its terminology.

\begin{figure}[t]
    \centering
    \includegraphics[width=1.0\columnwidth]{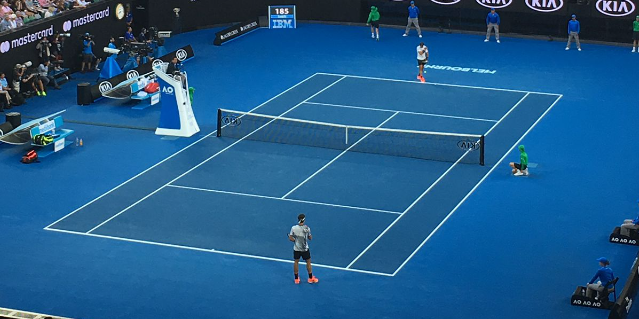}
    \caption{Photo of a Tennis Match}
    \label{fig:tennis}
\end{figure}

Tennis is a game played between two players that involves them hitting the ball to the other side of the court. \Cref{fig:tennis} is showing a photo of a regular tennis court. A \textit{rally} is the shot sequence containing all the shots starting with the serve of one player until a point is won or lost. Points can be won through opponent errors or successful shots that the opposing player was not able to reach. An error occurs, if the ball lands outside of the court or in the net. A \textit{fault}, e.g. by hitting the ball into the net on the first serve allows a second attempt; a \textit{double fault} gives the opponent the point. 

A detailed explanation of the scoring system in Tennis is not necessary for this paper. However, it is important to note, that a match consists of points, games, and sets. Scoring sufficient points earns a player a game; accumulating enough games secures a set, and claiming enough sets leads to a match victory. Each game starts with a serve from the right side of the court, called the deuce side, and the ball has to land in the opponent's left serving box. The next point then starts with a serve from the left side of the court, referred to as the advantage side and the serve has to land in the opponent's right service box. The service boxes mark specific court areas where serves must land, and players take turns serving between each game within a set. More details about the scoring system and rules in tennis can be found at \textit{USTA}~\cite{usta-tennis-scoring}.


\section{Match Point AI}
\label{sec:framework}

\begin{figure}[t]
    \centering
    \includegraphics[clip, trim=0cm 0cm 0cm 0.6cm, width=0.842\columnwidth]{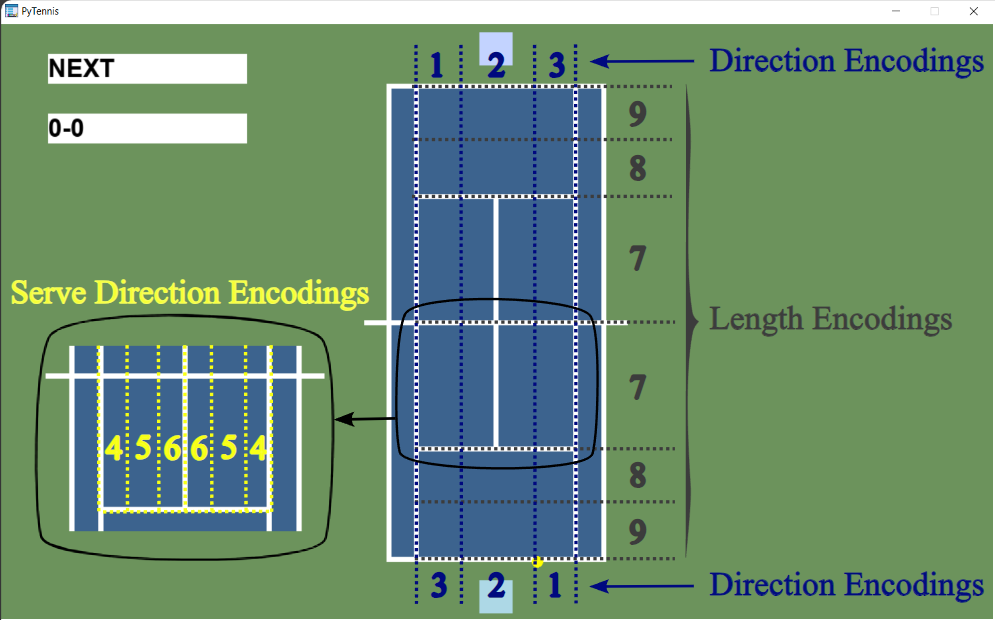}
    \caption{Tennis Shot Encoding}
    \label{fig:shot_encodings}
\end{figure}

In a real-world tennis match, many factors can influence the course of a match and the outcome of individual points. There can be external factors such as rain delays or the support by fans. On the court, there can be foot faults when serving, time violations, and medical timeouts. The outcome of a rally is heavily influenced by the shots in that rally, particularly the ball velocity, the player's movements and position on the court, and the direction of the previous shot. To tackle the shot direction selection problem in tennis, we model tennis as a non-deterministic game. Each action consists of an active choice of direction and a probability for that shot to be a winner or an error. 
For the shot direction encoding, we differentiate between serves and normal shots. In \Cref{fig:shot_encodings} we illustrate how the serve and shot directions are encoded.
This encoding is used in the dataset from the match charting project \cite{data_set}. Because we used this dataset to extract the error and winner probabilities for the individual shots, we made use of the same shot encoding.
The probabilities for an error or a winner are different depending on the current game state. A game state in \textit{Match Point AI} includes the direction of the opponent's previous shot, it considers which player was opening the rally with a serve and whether that serve \mbox{was a first or a second serve.} We neglect the players' positions and movement directions in the game states because no sufficient real-world data about this is available.

Furthermore, in \textit{Match Point AI}, the players follow the typical tennis match rules (see section II). 
When it is one of the player's turns to choose an action in an ongoing rally, he can choose to play the ball in the three different directions, encoded as \textit{1}, \textit{2}, and \textit{3}. If it is a player's turn to serve, he can choose between the action encodings \textit{4}, \textit{5}, and \textit{6}. The shot length encoding is used for visualization purposes. It is based on the match charting project dataset as well. Still, it does not influence the winner and error probabilities of individual shots and the depth can not actively be chosen by the players in \textit{Match Point AI}.

To summarize, each shot in \textit{Match Point AI} consists of an active choice of direction of one of the players and a stochastic component, whose error and winner probabilities are dependent on the current game state. The probability of the return being an error is for example higher when the previous serve was a first serve instead of a second serve because the second serve is usually easier to return than a first serve. \\
The Source Code of Match Point AI can be found 
at \\\url{https://github.com/cnuebel98/Match\_Point\_AI\_Public.git}

\section{Data-Driven Bot Strategies}

From related work on game AI, it is known that the performance of an AI agent depends on the quality of its opponent model. Especially, in search-based methods like MCTS, in which the opponent model drives the simulations, it can make a difference in the final agent's performance~\cite{DocDoe2017}.

For this purpose, we derived data-driven bot behavior using data from the match charting project spanning from 2017 to 2023. It consists of shot-by-shot tennis data for 295,354 rallies of singles matches in professional men's tennis. It covers 30 attributes of each tracked rally, including the tournament at which the match was played, the two opposing players, the scores, and the directions and types of all shots in the rallies. 

While players of real-world tennis differentiate between shot types, those are not considered throughout Match Point AI. In general, special shots, such as slice, volley, stops, or lob only make up 20\% of the data set and are therefore omitted from the current version (see Section A in Supplementary Materials).

Based on the introduced dataset two different bot strategies were created. 
For this purpose, we differentiate between three main shot types: the serve, the return, and other shots.
For the \textit{Serve} we differentiate between the side of the court from which it is hit, whether it was a first or second serve, and in which of the three different directions of the service box it was hit. A total of 36 different probabilities are included for the bot's serve behavior.
For the \textit{Return} the direction and the corresponding error and winner probabilities are dependent on the previous serve. Given 12 serve scenarios, 3 directions, and 3 possible outcomes (error, winner, undecided), a total of 108 probabilities is extracted.
For the remaining \textit{Normal Shots}, we consider the player who initiated the serve (due to their impact on the remaining rally), if it was the first or second serve, and the player's possible directions and the shot's possible outcomes. Once again a total of 108 \mbox{probabilities is extracted.}

For creating different bots this analysis was done on different parts of the datasets. The \textit{Djokovic Bot} was created, only extracting probabilities from rallies in which Novak Djokovic was playing. 
The \textit{Average Bot} is based on all rallies in the specified time frame. The composition of the Average Bot can be seen in table \Cref{tab:Composition Average Bot} found in the supplementary materials.

\section{Experiments and Results}
\label{sec:experiments}

\textit{Match Point AI} enables us to conduct numerous match simulations with diverse settings, offering a wide range of possibilities for the exploration of tennis strategies and the performance analysis of different MCTS agents.
Three sets of experiments were conducted for this paper, to illustrate different ways of how \textit{Match Point AI} can be used.

In the first experiment, the Average Bot competed against the Djokovic Bot in 500 simulated matches. Out of these, the Djokovic Bot emerged victorious in 82.6\% of them. Comparatively, between 2017 and 2023, the real-world Novak Djokovic participated in 394 competitive matches, winning 336 of them, resulting in a match-win rate of 85.3\%. While there is a difference of 2.7\%, it's important to acknowledge the factors of real-world tennis, which are not modeled in the simulations of \textit{Match Point AI}. The comparison shows that Match Point AI is already able to produce similar results to the ones observed in real games and may therefore be used to produce new and interesting insights into the shot direction selection problem in tennis.

With the second set of experiments, we want to compare different characteristics found in the generated data in Match Point AI to real-world data and assess the effectiveness of various selection policies in the MCTS algorithm. Therefore, we simulate matches where MCTS agents compete against two bot strategies.
The MCTS agents using different policies and their match and point win rates against the Average and the Djokovic Bot can be seen in \Cref{tab:overall_results_200_matches}. It illustrates the correlation between point win rates and match win rates. Winning even slightly more points than the opponent leads to substantially higher match-win rates. This correlation aligns with real-world data. \Cref{tab:player-stats} displays several professional tennis players' point and match win rates and the correlation found in our generated data can be seen in this real-world data as well.

\begin{table}[t]
    \centering
    \caption{Win Percentages of MCTS Agents against the Djokovic and the Average Bot - Win Rates of MCTS Agents out of 200 Matches}
    \label{tab:overall_results_200_matches}
    \begin{tabular}{|c|c|c|}
        \hline
        \textbf{MCTS Agents} 
        & \multicolumn{2}{c|}{\centering\textbf{Average Bot}} \\ 
        \cline{2-3}
        \textbf{Selection Policy} & \textbf{Point-Win Rate [\%]} & \textbf{Match-Win Rate [\%]} \\
        \hline
        \specialrule{0.07em}{0em}{0.1em}
        \specialrule{0.07em}{0em}{0em}
        UCT & 51.99 & \textbf{71.00}\\
        \hline
        Random & \textbf{52.14} & 69.50\\
        \hline
        Greedy & 51.58 & 66.00\\
        \hline
    \end{tabular}

\vspace{0.5em}

    \begin{tabular}{|c|c|c|}
        \hline
        \textbf{MCTS Agents}  
        & \multicolumn{2}{c|}{\centering\textbf{Djokovic Bot}} \\
        \cline{2-3}
        \textbf{Selection Policy} & \textbf{Point-Win Rate [\%]} & \textbf{Match-Win Rate [\%]} \\
        \hline
        \specialrule{0.07em}{0em}{0.1em}
        \specialrule{0.07em}{0em}{0em}
        UCT & \textbf{49.65} & \textbf{46.50}\\
        \hline
        Random & 49.03 & 39.00\\
        \hline
        Greedy & 49.34 & 41.50\\
        \hline
    \end{tabular}
\end{table}

\begin{table}[t]
    \centering
    \caption{Real-World Player Statistics}
    \label{tab:player-stats}
    \begin{tabular}{|c|c|c|c|}
        \hline
        \textbf{Player} & {\textbf{Ranking}} & {\textbf{Point Wins [\%]}} & {\textbf{Match Wins [\%]}} \\
        \specialrule{0.07em}{0em}{0.1em}
        \specialrule{0.07em}{0em}{0em}
        N. Djokovic & 1 & \textbf{54.86} & \textbf{85.86} \\
        \hline
        D. Medvedev & 3 & 52.71 & 74.25 \\
        \hline
        A. Zverev & 6 & 52.57 & 72.03 \\
        \hline
        C. Ruud & 11 & 50.57 & 64.67 \\
        \hline
        J. Struff & 21 & 49.86 & 49.52 \\
        \hline
    \end{tabular}
\end{table}

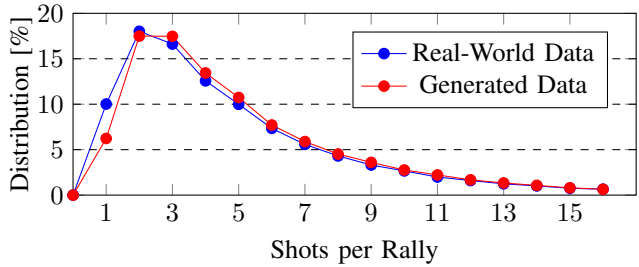
\begin{figure}[t]
    \centering
        \begin{tikzpicture}
            \begin{axis}[
                width=0.5\textwidth, 
                height=4cm, 
                xlabel={Shots per Rally},
                ylabel={Distribution [\%]},
                ymin=0,
                ymax=20,
                xmin=0,
                xmax=17,
                xtick={1,3,...,15},
                legend style={at={(0.95,0.9)},anchor=north east},
                ylabel style={yshift=-5pt},
                legend entries={Real-World Data, Generated Data},
            ]
                \addplot[blue, mark=*] coordinates {(0,0) (1,10.02) (2,18.02) (3,16.62) (4,12.57) (5,9.99) (6,7.34) (7,5.58) (8,4.31) (9,3.31) (10,2.66) (11,1.99) (12,1.6) (13,1.24) (14,1.01) (15,0.75) (16,0.65)};
                \addplot[red, mark=*] coordinates {(0,0) (1,6.23) (2,17.5) (3,17.47) (4,13.44) (5,10.74) (6,7.71) (7,5.89) (8,4.52) (9,3.6) (10,2.76) (11,2.22) (12,1.68) (13,1.33) (14,1.07) (15,0.8) (16,0.64)};
                \draw[dashed] (axis cs:0,5) -- (axis cs:17,5);
                \draw[dashed] (axis cs:0,10) -- (axis cs:17,10);
                \draw[dashed] (axis cs:0,15) -- (axis cs:17,15);
            \end{axis}
        \end{tikzpicture}
    \caption{Comparison of Rally Length Distributions}
    \label{fig:rally_length_dist}
\end{figure}

Furthermore, the rally length distribution of the rallies generated in \textit{Match Point AI} was compared to the distribution in the real-world data. While no significant difference was found between the total distributions, we can see in \ref{fig:rally_length_dist}, that the real-world data contains 3.79\% more rallies with only one shot, compared to the generated data. A one-shot rally only consists of a first-serve winner, an ace. This means it is harder to hit an ace in Match Point AI compared to real-world tennis. This could be due to the simplicity, with which tennis is modeled in Match Point AI, compared to the \mbox{complexity of real-world tennis.}

We then compared the effectiveness of different selection policies in the MCTS algorithm when used in \textit{Match Point AI}. In all matches, agents were using a \textit{Greedy} decision policy.
For the different selection policies, we see that slight variations in the point-win rate can result in drastic differences in the match-win rate.
Overall, in both matchups agents using the \textit{UCT} selection policy were performing best.

\begin{figure}[t]
    \centering
    \includegraphics[width=0.47\textwidth]{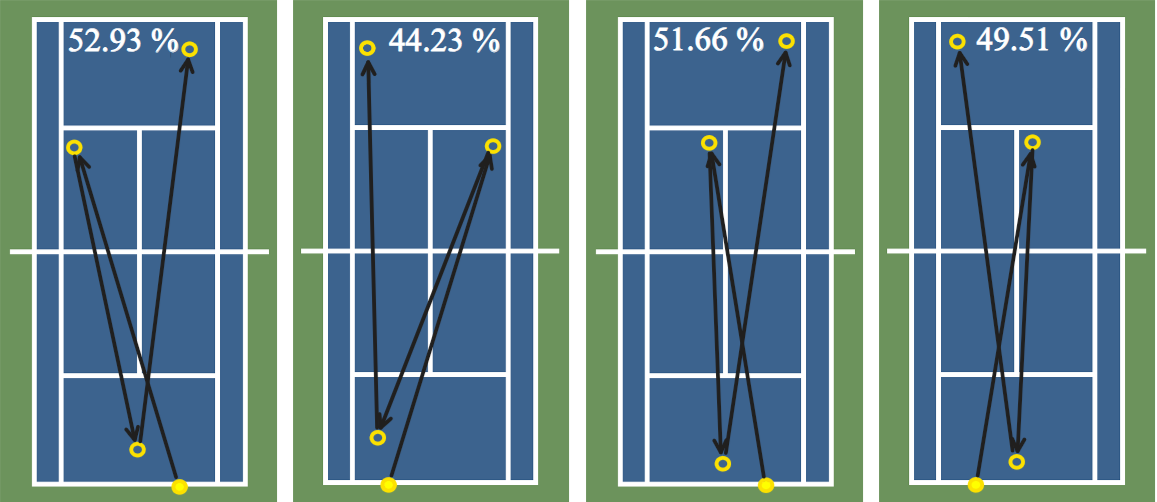}
    \caption{Most Frequent Shot Patterns and their Point Win Rates (left to right):
    First Serve from the Deuce Side, First Serve from the Advantage Side, Second
    Serve from the Deuce Side, Second Serve from the Advantage Side}
    \label{fig:shot_patterns}
\end{figure}

From this data, we also extracted the most frequent shot direction selection strategies in specific situations in a tennis rally used by the MCTS agents against the Djokovic Bot. The results are visualized in \ref{fig:shot_patterns}. We only look at the first three shots in a rally when the agents are serving first and also second serves from both the advantage and the deuce side of the court.
An interesting observation is that the agents are trying to make the Bot run. On the far left court for example, after placing the first serve to the far left side of the opponent's service box, the third shot is placed to the far right corner of the court. The same can be seen for the first serve from the advantage side on the inner left court. Making your opponent run is a common strategy in tennis. That agents can learn this strategy does however speak to the ability of Match Point AI to generate realistic tennis data as well as to the MCTS algorithm capability to solve the shot direction selection problem.

Finally, to analyze the effect that different parameter settings can have on the performance of the MCTS algorithm in \textit{Match Point AI}, in the third experiment we simulate matches with the MCTS agent using \textit{UCT} as selection and \textit{Greedy} as decision policy. 
We adapt this agent by using three different $C$ values ($\sqrt{2}-0.5$, $\sqrt{2}$ and $\sqrt{2}+0.5$) during the simulation phase and let these adapted agents play against the Average Bot in 100 matches each. The results in \Cref{tab:c-values} show, that the agent using a $C$ value of $\sqrt{2}+0.5$ wins significantly more points against the Average Bot compared to agents using the other two $C$ values.

\section{Conclusion and Future Work}

In this work, we present the tennis match simulation environment, \textit{Match Point AI}, in which tennis is modeled as a non-deterministic game. Further, we showcase how different data-driven bot strategies, reflecting real-world player behaviors, can engage in matches against different MCTS-based agents. By comparing the shot-by-shot data generated by \textit{Match Point AI} with real-world tennis data, we observe realistic parallels, such as the rally length distributions and correlations between point and match win rates. By analyzing the strategies devised by the agents, we demonstrate how MCTS can be used in \textit{Match Point AI} to optimize the shot direction selection problem in tennis. Here, different policies and parameter settings are used in the MCTS algorithm to determine, which MCTS adaptation is best suited to solve this problem.

Even though the first experiments conducted with \textit{Match Point AI} show promising results, there are several opportunities for future improvements of the framework. Most of these improvements, however, largely depend on incorporating more and better tennis data. 
The dataset used in this paper is a valuable source of real-world shot-by-shot tennis data, with untapped potential for example for adding shot types like drop shots, lobs, and volleys. However, it lacks detailed information on ball velocities and player positions and their movements, which are crucial for understanding a player's shot direction decisions.

There are already other ways of tracking tennis data, which can keep track of the data automatically and more precisely than the crowd-sourced match charting project, such as electronic line calling with \textit{Hawk-Eye}~\cite{hawkeye,mecheri2016serve}.
The \textit{Hawk-Eye} technology nowadays replaces line judges at major tournaments and tracks ball placement down to the millimeter.
Including data like this would drastically increase the capabilities of \textit{Match Point AI} and the bot strategies would represent the real-world player's behavior much more precisely.

\begin{table}[t]
    \centering
    \caption{Win Percentages of MCTS UCT/Greedy Agent with Parameter
Adaptations against Average Bot}
    \label{tab:c-values}
    \begin{tabular}{|c|c|c|}
        \hline
        \textbf{MCTS Agents} 
        & \multicolumn{2}{c|}{\centering\textbf{Average Bot}} \\ 
        \hline
        \multirow{2}{*}{$C$ Value} & 
        Point-Win & Match-Win  \\ 
         & Rate [\%] & Rate [\%] \\
        \specialrule{0.07em}{0em}{0.1em}
        \specialrule{0.07em}{0em}{0em}
        $\sqrt{2} - 0.5$ & 50.53 & 55.00\\\hline
        $\sqrt{2}$ & 51.71 & 63.00\\\hline
        $\sqrt{2} + 0.5$ & 52.26 & 75.00\\
        \hline
    \end{tabular}
\end{table}

It would also be interesting to see, if the agents come up with personalized strategies against different Bots, mirroring other player's behavior than that of Novak Djokovic, because the best strategy to win against Djokovic is probably not the best strategy to win against Rafael Nadal.
The shot pattern analysis was conducted for a few very specific scenarios in a tennis match. Tennis and also the resulting data of \textit{Match Point AI} offers a wide range of other scenarios, that would be interesting to explore as well. 

Finally, in this paper, we used the MCTS algorithm and analyzed the influence of different policies and different parameter settings on its performance in winning points in \textit{Match Point AI}. Despite our attempts to tune the agent's parameters, no agent has been capable of defeating the Djokovic bot. Therefore, future testing will involve the development of further MCTS variants and other agents to find innovative solutions to the shot direction selection problem in tennis.

\bibliographystyle{unsrturl}
\bibliography{references.bib}

\newpage
\renewcommand{\appendixname}{Supplementary Material}
\appendix

\section*{A - Distribution of Shot Types}

\begin{figure}[h]
        \centering
        \begin{tikzpicture}
            \begin{axis}[
                ybar,
                ylabel={Distribution [\%]},
                xtick=data,
                xticklabels={Normal shots, Slice, Volley, Drop shots, Lobs, Others},
                xticklabel style={rotate=30, anchor=east},
                ymin=0,
                ymax=100,
                width=8.5cm,
                height=4cm,
                bar width=0.7cm,
                nodes near coords,
                nodes near coords align={vertical},
                grid=both,
                ]
                \addplot coordinates {
                    (1, 80.77)
                    (2, 12.41)
                    (3, 2.6)
                    (4, 1.41)
                    (5, 1.39)
                    (6, 1.42)
                };
            \end{axis}
        \end{tikzpicture}
    \caption{Shot Type Distribution of the Real-World Dataset}
    \label{fig:shot_types_real_world}
    \end{figure}
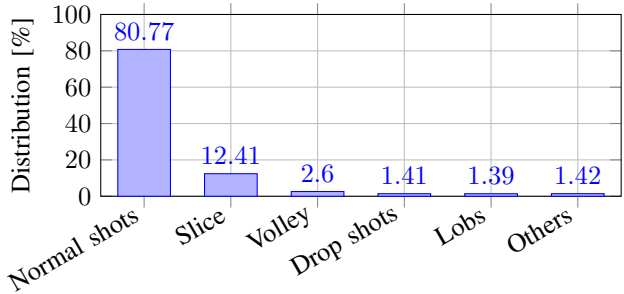

When analyzing the shot types documented in the dataset illustrated in Figure \ref{fig:shot_types_real_world}, it is apparent that real-world players rarely utilize the shot types we neglect, accounting for less than 20\% of occurrences.

\section*{B - Composition of the Average Bot Strategy}

Out of the 420 tennis players involved in the rallies analyzed for the Average Bots' behavior, nearly half of the Bots' behavior was attributed to just 10 players. This disproportionate influence stemmed from the extensive tracking of rallies involving these players by the match charting community. The following table shows the share of the 10 most included players.

\begin{table}[h]
    \centering
    \caption{Composition of the Average Bot}
    \label{tab:Composition Average Bot}
    \begin{tabular}{|c|c|}
        \hline
        \multirow{2}{*}{\textbf{Player}} & \textbf{Share in Average} \\
        & \textbf{Bot's Behaviour [\%]}\\
        \specialrule{0.07em}{0em}{0.1em}
        \specialrule{0.07em}{0em}{0em} 
        Roger Federer & 6.7\\
        \hline
        Novak Djokovic & 6.3\\
        \hline
        Daniil Medvedev & 6.2\\
        \hline
        Rafael Nadal & 5.4\\
        \hline
        Stefanos Tsitsipas & 5.0\\
        \hline
        Dominic Thiem & 4.6\\
        \hline
        Alexander Zverev & 4.1\\
        \hline
        Andrey Rublev & 3.8\\
        \hline
        Jannik Sinner & 3.3\\
        \hline
        Gael Monfils & 3.1\\
        \hline
        Others & 51.5\\
        \hline
    \end{tabular}

\end{table}


\end{document}